\documentclass[twocolumn,aps, nofootinbib, showpacs,showkeys]{revtex4}

\usepackage{amsmath}
\usepackage{graphicx}
\usepackage{dcolumn}
\usepackage{bm}

\begin{document}

\title[]{The role of grammar in transition-probabilities of subsequent words in English text}

\author{Rudolf Hanel$^1,2$; Stefan Thurner$^{1,2,3,4*}$;}

\affiliation{$^1$Section of Science of Complex Systems; Medical University of Vienna; Spitalgasse 23; A-1090; Austria; \\ 
$^2$Complexity Science Hub Vienna, Josefst\"adterstrasse 39, 1080 Vienna, Austria; \\
$^3$Santa Fe Institute; 1399 Hyde Park Road; Santa Fe; NM 87501; USA;\\
$^4$IIASA, Schlossplatz 1, 2361 Laxenburg, Austria;}
\email{stefan.thurner@meduniwien.ac.at}

\begin{abstract}
Sentence formation is a highly structured, history-dependent, and sample-space reducing (SSR) process. 
While the first word in a sentence can be chosen from the entire vocabulary, typically, the freedom of 
choosing subsequent words gets more and more constrained by grammar and context, as the sentence progresses. 
This sample-space reducing property offers a natural explanation of Zipf's law in word frequencies, however, it fails 
to capture the structure of the word-to-word transition probability matrices of English text.  
Here we adopt the view that grammatical constraints (such as subject--predicate--object) 
locally re-order the word order in sentences that are sampled with a SSR word generation process. 
We demonstrate that superimposing grammatical structure--as a local word re-ordering 
(permutation) process--on a sample-space reducing process is sufficient to explain both, 
word frequencies and word-to-word transition probabilities. 
We compare the quality of the grammatically ordered SSR model in reproducing several test statistics of 
real texts with other text generation models, such as 
the Bernoulli model, the Simon model, and the Monkey typewriting model.
 


\end{abstract}

\maketitle

\section{Introduction}
\begin{figure}[t]
\centering
\includegraphics[width=0.9\columnwidth]{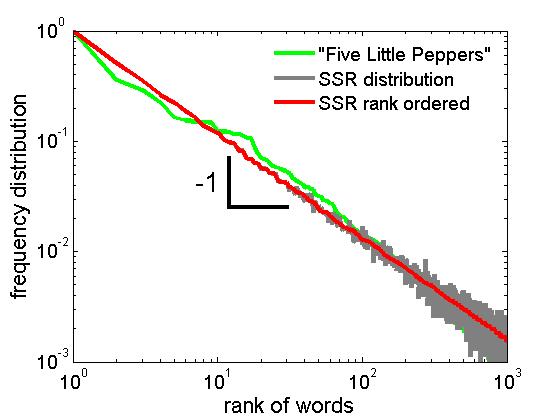}
\caption{Rank ordered frequency distribution of the most frequent 1000 words in the novel 
``Five little peppers and how they grew'', by M. Sidney (green). 
The result from the SSR model (gray) and its rank ordered frequency distribution (red) 
show an exact Zipf's law, and are both invariant under grammatical re-shuffling.
The SSR sequence has been produced for approximately $W=5000$ words, and $N=50,000$ samples (gray). 
} 
\label{fig1}
\end{figure}

After almost a century of work, understanding statistical regularities in language is still work in progress. 
Maybe the most striking statistical feature is that rank ordered distributions of word frequencies follow an approximate
power law, 
\begin{equation}
	f(r) \sim r^{-\alpha} \quad,  
\end{equation} 
where $r$ is the rank assigned to every word in a given text; the most frequent word has rank one, the second most 
frequent has rank two, etc. For most word-based texts, one finds $\alpha \sim 1$, independent of language,
genre, and time of writing. This ``universal'' feature is called Zipf's law \cite{zipf72}. Figure \ref{fig1}  shows the rank 
distribution of words in the novel ``The Two Captains'' by  H.K.F. de la Motte Fouqu\'e (green). 

There are several ways to understand Zipf's law through entirely different mechanisms. 
Zipf's first qualitative explanation of the phenomenon was based on communication ``efforts'' of sender and receiver 
\cite{zipf72}, an idea that was later expressed in an information-theoretic framework 
\cite{mandelbrot1953,Harremoes:2001,sole2003,Corominas-Murtra:2011}. 
The first quantitative linguistic model by H. Simon features the idea of preferential attachment, i.e.,  
words are added to the text with a probability that is proportional to their previous appearance in the text. 
New words are added at a low, constant rate \cite{simon:1955}. Zipf's law follows immediately from those two 
assumptions. Preferential attachment models were later refined \cite{zanette2005}, relating Zipf's law to Heap's law \cite{Heaps1978}. 
The conceptually simplest way to understand Zipf's law are random typewriting models, 
where words and texts are created by randomly typing on a typewriter \cite{li1992,miller1957,miller1963}.
Yet another route to Zipf's law was introduced on the basis of sample-space reducing (SSR) 
processes \cite{Language1-MHT}, which successively reduce their sample-space (range of potential outcomes) as they  
unfold \cite{Zipf1C-MHT}. SSR processes generically produce power laws, and Zipf's law in particular \cite{driving}. 
Think of how sentences are formed: one can pick any word to start a sentence. Once the first word is chosen, 
grammar and context constrain the possibilities for choosing the second word. 
The choice of the second word further constrains the possibilities for the third word, and so on; 
The sample space of possible words
generally reduces as the sentence forms. 
In this view of sentence formation, grammar and context constrain the choice of words later in the sentence; 
therefore text generation is a SSR process, and Zipf's law must follow. 
The existence of grammatical and contextual constraints allow us--at the receiving part of a communication--to 
complete sentences in advance, and to anticipate words that will appear later. 
This (at least partially) ordered hierarchical structure {\em guides} sentence formation and allows a receiver to robustly decode messages \cite{Partee:1976}.  

\begin{figure}[t]
\centering
\includegraphics[width=0.9\columnwidth]{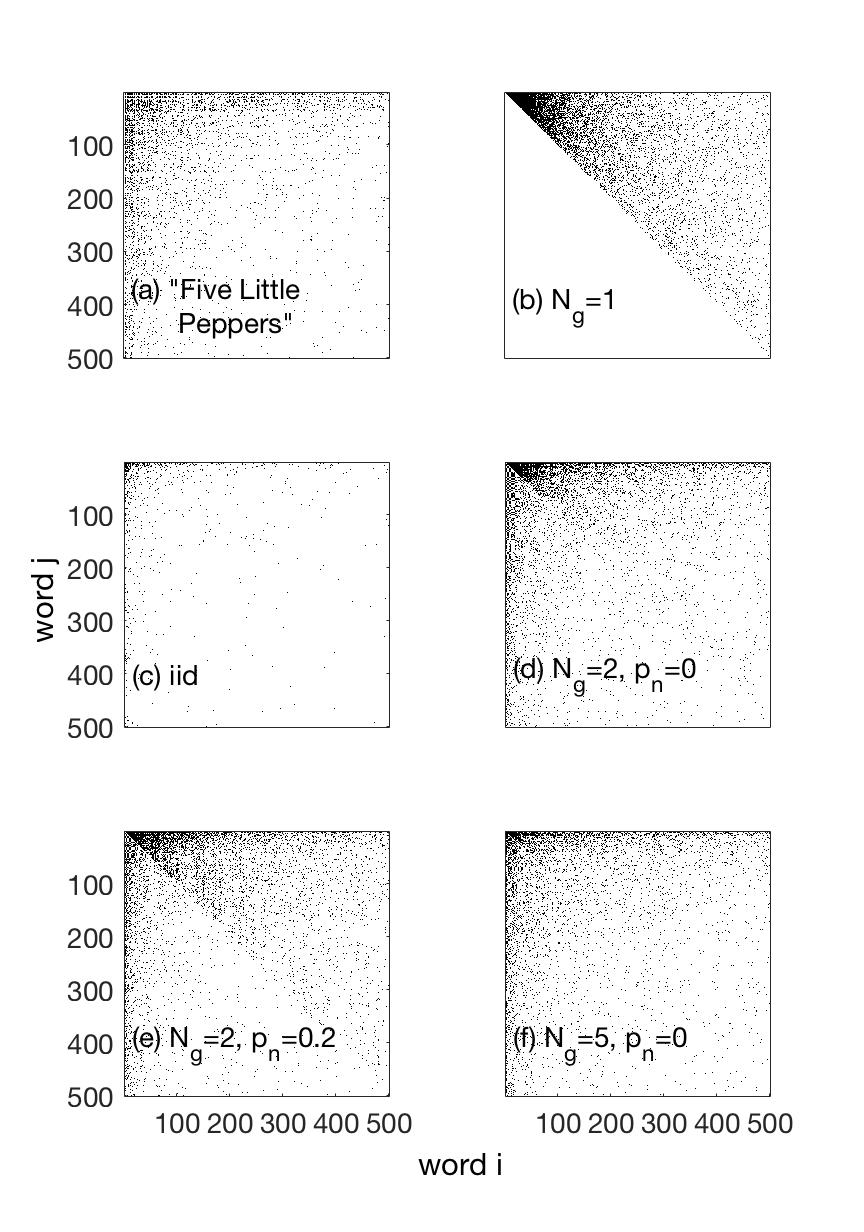}
\caption{
(a) Word transition matrix $A_{ij}$ (first index is x-axis, second is y-axis) of the novel 
``Five little peppers and how they grew'' means that word $i$ followed word $j$ at least once in the text. 
Words are ordered along the axis according to their frequency of appearance. 
(b) Transition matrix of a pure SSR process. States are ordered according to their natural index $i=1,\cdots,W$.
The triangular structure is visible. 
(c) The Bernoulli model for the same novel shows the matrix for words sampled independently from the marginal word frequency distribution of the novel.
(d-f) Transition matrices for goSSR processes for $N_g=2$ and $5$ grammatical categories. 
In (d) and (e) we compare the situation with different probabilities for neutral elements $p_{n}=0$ and $p_{n}=0.2$.  
As $N_g$ increases, the goSSR transition matrices begin to resemble those of actual text.
All transition probabilities are computed for a lexicon containing 
roughly $4,800$ words. Samples consist of approximately $88,000$ word transitions and
$5,200$ sentences. 
}
\label{fig2}
\end{figure}

To understand Zipf's law of word frequencies, however, is not the end of the story. 
The word frequency distribution can be seen as the marginal distribution of the word transition probabilities, $p(i|j)$, 
the probability to produce word $i$, given that the previous word was $j$. 
Figure  \ref{fig2}a shows the transition matrix for ``Five Little Peppers''. 
Pure SSR processes show a triangular structure in the transition probabilities, see 
Fig. \ref{fig2}b, which obviously does not match the empirically observed transition matrices of real texts. 
Empirical transition probabilities look similar to transition matrices 
that correspond to independently sampled word sequences
(Bernoulli model) with the correct 
marginal
empirical word distribution functions, Fig. \ref{fig2}c.
Empirical transition probabilities look as if they were random (in this sense), 
even though text generation is obviously a highly structured generative processes.

In this paper we assume that the formation of word sequences (sentences) is a combination of two processes: 
The {\em word selection process}, selects the words that are needed to encode a narrative or to convey a coherent message 
or meaning. 
The other process is {\em grammar}, which brings the selected words into a specific order.  
We assume that the word selection process is of SSR type: 
the context created by the generated words restricts the usage of other words as the sentence progresses. 
The SSR structure of word selection becomes plausible by realizing that any story line needs to connect 
the ``protagonists'' with their context. If we think of context as a network of words, 
then the process of connecting any chosen word to a given protagonist is comparable to a {\em targeted diffusion} 
process on the word network, where protagonists (and other central words) are the targets in this process. 
Targeted diffusion is an example of SSR processes, and leads to generic power laws in visiting frequencies \cite{targdiff}.
Grammatical ordering can be thought of as a post-processing of the word stream 
generated by the word selection process. 
It establishes a local word order, which differs from the order the words were generated, 
but is in line with grammatical expectations, such as the subject--predicate--object (SPO) 
order that is typically used in English sentences. 
In other words, grammar locally scrambles (permutes) the order of the word selection stream. 

We will show here that if the word selection process is a pure SSR process, 
and if generic grammatical rules locally re-shuffle the word order of the word selection stream, 
the resulting word transition probabilities have statistical properties that closely resemble 
the empirically observed ones. 
We will see that grammar strongly masks the triangular structure introduced by the 
word selection process. This implies that empirical word transition probabilities provide us 
with limited information about the underlying word selection process that generates an 
information-carrying narrative. 
In particular, with a simple SSR model of word generation we demonstrate how imposing grammatical rules of variable strength
changes the transition probabilities from structured 
(triangular, Fig. \ref{fig2}b) to seemingly unstructured, Fig. \ref{fig2}c.

We first discuss SSR processes and their frequency distributions and then introduce a modification, 
where the local word order of SSR sequences is permuted to conform with a grammatical word order, such as SPO.
To implement a
``grammatically ordered'' SSR  (goSSR) process we assign grammatical labels 
$c\in\{1,\cdots,N_g\}$ to all words contained in the lexicon of used words. 
The labels determine the {\em local} order in which words appear in a sentence.
We compute statistical properties of natural language in English text corpora \cite{gutenberg} 
and compare them with those obtained from goSSR processes. 
We find that three to five grammatical labels are sufficient to produce realistic results. 
We finally compare the Simon-, the random typewriting-, and the goSSR-models for 
specific English texts with respect to various statistical measures that can be used as 
a test-statistic for hypotheses testing.
As a null-hypothesis we assume that sequences have been generated by a Bernoulli process. 
This provides us with a first quantitative understanding of how informative these models are with 
respect to actual text generation.

\subsection{Grammatically ordered SSR model}

We first discuss the sample-space reducing  process for sentence formation and then augment it with a toy grammar.
SSR processes are characterized by $W$ linearly ordered states. 
For example think of a staircase. The lowest stair is state 1, the next step is state 2, and so on; 
at the top of the staircase we have state $W$. 
Imagine a ball bouncing downward this staircase with random jump sizes. We begin at state $W$.   
The ball can jump to any of the $W-1$ lower states; 
say it jumps to state $x_1$ (subscript indicates the first jump).  Obviously, $1\leq x_1\leq W-1$. 
Again, the next state, $x_2$, can only be a lower state, $1\leq x_2< x_1$. 
After a sequence of $n-1$ visits to states  $x_1,\dots,x_{n-1}$, the ball reaches the bottom 
of the staircase, $x_{n}=1$. 
At this state the process needs to be restarted, which means lifting the ball to any randomly chosen state, 
$1<x_{n+1}\leq W$. If the process gets restarted multiple times, the visiting distribution of the process 
to states $i$ appears to be exactly Zipf's law,
\begin{equation}
p_i =  \frac{1}{Z} \frac{1}{i} \quad ,
\label{SSRdist}
\end{equation}
where $Z$ is a normalization constant \cite{Zipf1C-MHT}.

Sentence formation can be seen as an SSR process \cite{Language1-MHT}. 
Words are not randomly drawn from the sample space of all possible words (lexicon), 
but are used in context and grammatical order. 
The fact that words in a sentence restrict the usage of consecutive words, generates a SSR process. 
Imagine the first word in a sentence is randomly drawn from the entire lexicon with $W$ words (states), say ``The wolf". 
As soon as it is decided, the second word must be a verb (grammatical restriction), 
and it has to create a meaningful context (context dependent restriction) with the wolf, so we can not use ``typewrite". 
As the sentence progresses, typically more and more constraints occur and restrict word usage more and more.
Once the final word 
(the target of a sentence, state 1)  is reached  the process gets restarted; (the next sentence or subordinate clause starts)
Sample space reduction in text formation is necessary to robustly convey meaningful information, 
a fact that is for example exploited by text-completion apps. 

While rank ordered word frequency distribution functions can be explained by the hypothesis that sentence and 
text formation follows SSR processes, word transition probabilities can not.
Transition probabilities for pure SSR processes are triangular, see Fig. \ref{fig2}b, and do not resemble those of 
actual text, Fig. \ref{fig2}a.


We assume that the word selection process for a given narrative can be approximated by a SSR process 
that models the context-dependent restrictions only. 
Grammar enters as a process that enforces locally ``correct" word order. 
Effectively, it locally re-assembles the word sequences generated by the SSR word selection process, 
and destroys local correlations of word occurrences as generated by the SSR sequence. 
We now augment SSR models with a ``grammar'' that determines the local ordering of words.

To implement a toy-grammar, assume that there exist $N_g$ grammatical labels that are associated to words. 
Every word, $i=1,\cdots,W$ in the lexicon, carries one of the $N_g$ distinct labels, $L(i)\in\{1,2,\cdots,N_g\}$. 
For example, if $N_g=3$, the three labels could represent $S$, $P$, and $O$. 
For simplicity we assume that each label appears with approximately the same frequency in the lexicon.
In addition there exists a grammatical label $L(i)=0$ that we call the ``neutral'' label. 
Neutral elements are combined with the next non-neutral word (with label $L$) that follows in the text.
This word complex is then treated as a single word with label $L$.
The probability of finding neutral words in the text is denoted by $p_n$. 
We can now formulate the ``grammar rules'': 
\begin{itemize}
	\item (i) Words must follow a strict repeating pattern of grammatical labels, 
	$1\to 2\to 3\to \cdots\to N_g\to 1\to 2 \to  \cdots$.
	For example, if $N_g=3$ we will produce sentences with a sequence of labels: $\cdots\to S\to P\to O\to S\to P\to O\to\cdots$.	
	\item (ii) Missing labels are skipped. If $N_g=4$ and if label $3$ is not present in a particular sentence, but labels $1$, $2$ and $4$ are, 
	 we order words according to existing labels: $1\to2\to4$.
	\item (iii) Neutral elements (label $0$) do not change their relative position to the next non-neutral element following in the sentence. Neutral elements together with their adjacent non-neutral element form a complex that in grammatical reordering is treated as a single word. This means that the local SSR sequence order of grammatically neutral fragments is untouched by the grammatical ordering process. 
\end{itemize}
For example, an SSR sequence of words (states) $260, 120, 76, 45, 13, 12, 7, 1$ is generated in a $N_g=3$ 
grammar with the corresponding grammatical label sequence:
$3, 1, 0, 0, 2, 3, 0, 1$. The grammar-ordered label sequence is $1, 0, 0, 2, 3, 0, 1, 3$, and the 
grammar-ordered state sequence becomes
$120, 76, 45, 13, 260, 7, 1, 12$. 

The goSSR model of a given text is the following. Determine the number $W$ of distinct words in a text. 
Produce a random map $L$ that associates a label $0\leq L(i)\leq N_g$ with each word $i=1,2,\dots,W$. 
The neutral label $L(i)=0$ gets sampled with probability $p_n$, all other labels with probability $(1-p_n)/N_g$.
For each sentence in the text determine its length (in words) and generate a random SSR sequence of the same length.
Then re-order this SSR sequence according to the grammar rules to get a grammatically ordered sentence. 
We call this sampled new text, the {\em goSSR model} of some original text. It contains as many words 
and sentences as the original text. Each sentence is a goSSR sequence. 

For comparison, we consider three other models of text generation that yield power law distributed rank frequencies.
The simplest is produced by independently sampling words from the word frequency distribution of the original text, 
which  we refer to as the {\em Bernoulli model} of the original text. 
The others are the Simon model, and the random typewriting model of text, see Methods.

\begin{figure*}[ht]
	\centering
  \includegraphics[width=0.8\textwidth]{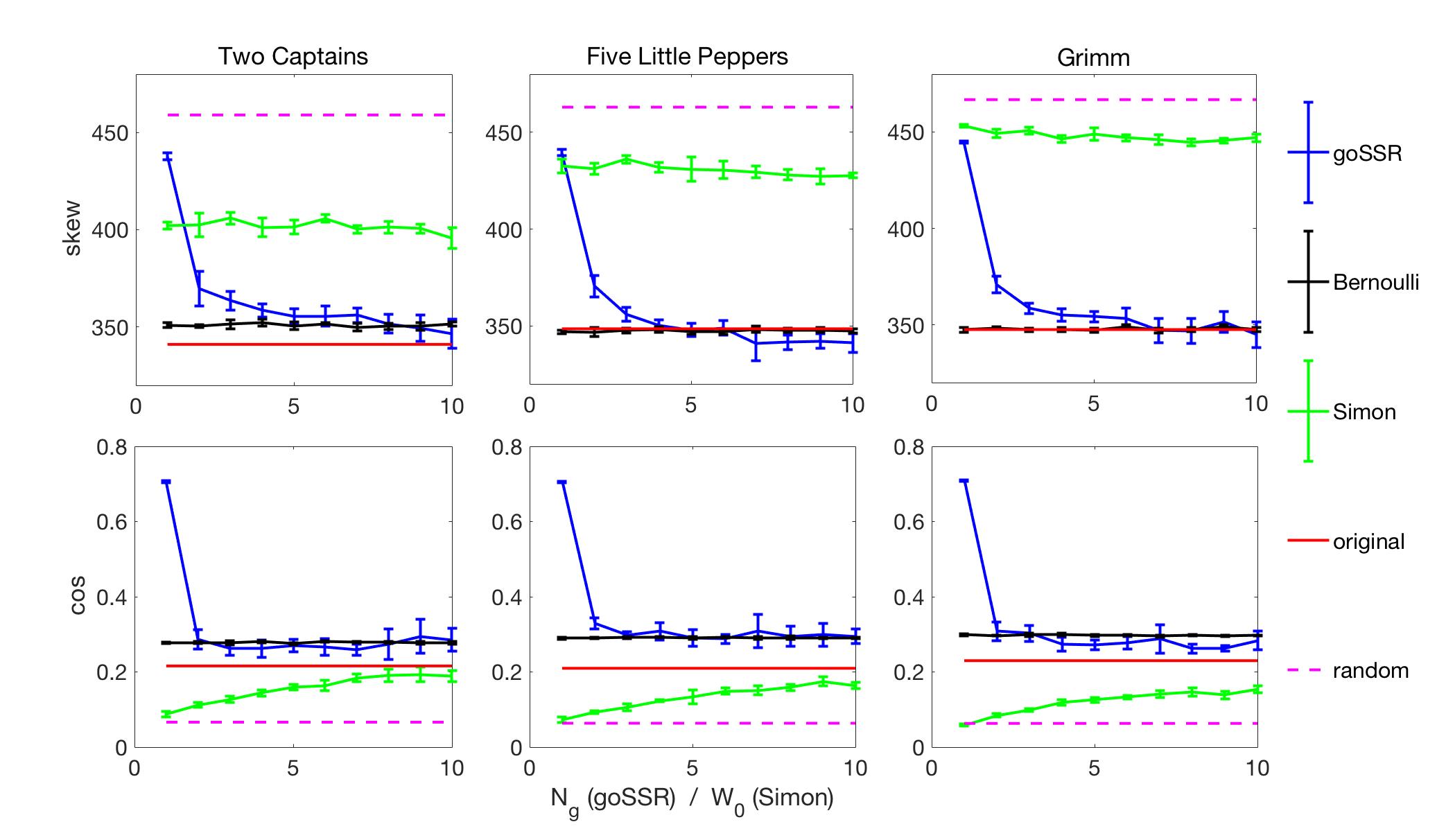}
  \caption{For three English novels two measures, the skew and a $\cos$-measure, are used to 
  compare the word transition matrices of the goSSR model (blue) those of real texts (red) and their 
  associated Bernoulli model (black). The values for the Simon model (green) with varying $W_0$ (from 1 to 10), 
  and the random typewriting model (magenta) for $V=30$, are shown. 
\label{fig4}	
	}
\end{figure*}

\section{Results}

We consider three English novels.
(i)  {\em The Two Captains} ({2cpns}), by F. de La Motte-Fouque, with a length of roughly 18,000 words, 
600 sentences, and a vocabulary of roughly 3,000 words. 
(ii)  {\em Five little peppers and how they grew} ({5lpep}), by M. Sidney,
with about 88,000 words, 5,200 sentences, and 4,800 distinct words. 
(iii) {\em Collected tales by the Brothers Grimm} ({grimm}), with 180,000 words, 
4,500 sentences, and about 4,800 words. 
For these texts we sample the corresponding goSSR- , Bernoulli-, Simon-, and the random typewriting-model with 
the same sentence length distributions as the corresponding original text. 
Since the vocabulary of the texts is too large for computing the Simon-, and random typewriting-models, we 
restrict ourselves to the word transitions between the $500$ most frequent words in the texts. 
For notation, we use $K_{ij}$ for the number of times we observe the word $i$ to follow word $j$
in a sentence of a text. $A_{ij}=K_{ij}/k_j$ estimates the conditional probability for $i$ to follow $j$, 
where $k_j$ is the number of times word $j$ appears in the text. 

We then compute two statistical measures 
that allow us to quantify properties of the transition matrices.
The {\em skew}  is a measure of the asymmetry of $A$, and the {$\cos$}-measure 
is a proxy for the largest eigenvalue of $A$.

We compute the transition matrices $A^{\rm text}$, $A^{\rm bernoulli}$, $A^{\rm goSSR}$, $A^{\rm simon}$, 
and $A^{\rm random}$, and present the corresponding measures in Fig. \ref{fig4}. 
The top line in Fig. \ref{fig4} shows the skew for 2cpns, 5lpep, and grimm. 
The bottom line shows the $\cos$-measure.
The values corresponding to the original texts are shown as red lines. 
The values for the Bernoulli models (black) are averages over 5 realizations of the model.
For the original text and the Bernoulli model the $x$-axis of the plot has no meaning 
since neither the original text nor the Bernoulli model depend on model parameters. 
For the random typewriting model (magenta) we fix the number of letters on the typewriter to $V=30$, see 
Methods. 
For the goSSR and the  Simon model the $x$-axis has a different meaning. 
For the goSSR model (blue) the $x$-axis is the number of grammatical labels $N_g=1,\cdots,12$, 
for the Simon model (green) it corresponds to the number of words, 
$W_0$ used for the initialization, see 
Methods. 
For all models we created $5$ realizations and present the mean and standard deviations. 
For the Bernoulli model the error bars are only slightly larger than the line-width. 
The goSSR model is computed with a neutral element probability of $p_n=0.1$. 
The value for the pure SSR models (no grammar, $N_g=1$) is marked with a black square.  

For the three novels we observe in Fig. \ref{fig4}  that for the two measures we find that the goSSR 
model outperforms both, the Simon-, and the random-typewriting models. 
For increasing $N_g$ the goSSR model approaches both, the skew and the $\cos$-measure of the real text.
For levels of  $N_g\sim3-5$ convergence is reached.
For the skew, both, the random typewriting and Simon model can not explain the real text value. 
For the $\cos$-measure the typewriting clearly fails. The Simon model approaches the real text value 
for large values of $W_0$.
The $\cos$-measure for the goSSR model is similar to the Bernoulli case, more so than the original text. 
There is almost no dependence on $N_g$, practically all values for $N_g>1$ are similar. 
Since typically the skew of written text is close or 
identical to that of the Bernoulli model (see, 5lpep and grimm), 
one may be inclined to interpret the skew measure as something like the ``grammatical depth'' of a text.  
However, keep in mind that the Bernoulli model can per se not explain the Zipf law in word frequencies;  
that is a massive exogenous input. 

\begin{table}
\caption{p-values for test statistics skew and $\cos$. 
The null hypothesis that the model texts 
have been generated by the Bernoulli model, at a $5$ \% confidence level must be rejected if 
$p<0.05$. 
Values for the goSSR model are shown for various $N_g$, the Simon model for $W_0=10$ and
the random typewriting model for $V=30$.
}
\begin{tabular}{l cc cc cc}
\hline
					& 2cpns  & 					& 5lpep  &  & grimm & \\ 
\hline 
					& skew 		& $\cos$ & skew 		& $\cos$ & skew 		& $\cos$ \\
\hline 
original 				& $0.0002$ 	& $0.0002$ & $0.59$& $0.0002$& $0.82$ & $0.0002$\\
Bernoulli 				& $0.55$		& $0.46$ & $0.57$ & $0.53$ &$0.51$ & $0.48$ \\
$N_g=1$ 		& $0.0002$	& $0.0002$& $0.0002$ & $0.0002$ & $0.0002$& $0.0002$\\
$N_g=3$		& $0.012$		& $0.009$ &$0.0002$ & $0.0002$& $0.0002$& $0.02$ \\
$N_g=5$		& $0.18$		& $0.07$&  $0.21$& $0.12$ & $0.025$& $0.0002$ \\
$N_g=7$		& $0.12$		& $0.116$& $0.41$ & $0.002$ & $0.0002$ & $0.0035$\\
$N_g=10$ 		& $0.27$		& $0.14$ & $0.019$& $0.12$& $0.079$& $0.0002$\\ 
Simon 	& $0.0002$ 	& $0.0002$& $0.0002$& $0.0002$& $0.0002$& $0.0002$\\
random    & $0.0002$	& $0.0002$& $0.0002$& $0.0002$& $0.0002$& $0.0002$\\
\end{tabular}
\end{table} 

In Table 1 we show the corresponding p-values. 
To compute those, consider the randomly generated texts (corresponding to a particular novel) 
by the different models as data. We would like to test whether we can reject the null hypothesis that this data has been 
generated by the Bernoulli model at a confidence level of $0.05$. 
To this end we sample the Bernoulli model of the particular novel for $5,000$ times, 
compute the respective measure for each realization of the model, rank the values, and compute the respective 
p-values for the $5$ realizations of models of the novel. 
If a value is smaller than the the smallest of the 5000 samples
drawn for the test statistics we interpret of the associated p-value as a number less than $1/5000=0.0002$.
We present the average of the 5 p-values for 
the three novels.
We find that  both measures for the original texts and their Bernoulli models are numerically similar, 
which confirms the similarity (on rough visual inspection) 
of transition matrices of original text and corresponding Bernoulli models, 
compare Fig.\ref{fig2}(a) and (b). 
The cases where we can not reject the null hypothesis are for the Bernoulli model itself, which for all test statistics 
is accepted with $p\sim 0.5$, and for the goSSR model for grammars with $N_g>4$, see Tab. 1.

Similarly, for the {\emph skew}-measure of ``Two Captains'', 
for $N_g\geq 4$ p-values also exceed the confidence level. That is, the transition probabilities of the 
goSSR process cannot be distinguished from the Bernoulli model with respect to the skew if the 
grammar becomes sufficiently complex.

Both measures indicate that original text in fact has transition matrices that resemble those of the corresponding 
Bernoulli model to a high degree, and that for reasonable choices of  $N_g$ and $p_n$, the goSSR model 
also matches these. The used measures indicate that the word transition matrices of goSSR 
models are located somewhere between real text and Bernoulli models in a statistical sense.  

We considered additional measures to derive test-statistics that allow us to 
compare the rank-increment distributions of the real texts and the models. 
These included the $L_1$-norm, the Kolmogorov-Smirnov distance, and the 
Kullback-Leibler divergence. 
They are not shown here since they add little additional aspects.
%
%
%
However, a note of caution for the naive interpretation of p-values is necessary. 
The skew and $\cos$-measure are among the most simple ones and are easy to interpret. 
There are many more, 
and it is conceivable that not all corresponding test-statistics necessarily confirm that the goSSR 
model outperforms the other models. 
In fact, we found that one test-statistic, based on  the Kolmogorov-Smirnov of rank-increment statistics, 
the Simon model performed slightly better than the corresponding goSSR model.
What is true for all considered test-statistics, however, is that goSSR models show 
values of test-statistics that are generally between real text and the Bernoulli model. 
Using different test-statistics for comparing models of complex phenomena 
may lead to distinct notions of similarity, which need not coincide. 
Different models may be adequate for some features (test-statistics) of the modeled phenomenon, 
while they may be quite inadequate with respect to others. 

\section{Discussion}

The main objective of this paper was to demonstrate that goSSR models can explain both, the shape of the marginal distribution function, i.e. Zipf's law, and 
important structural/statistical features
of the empirical word transition matrix, as shown in Fig. \ref{fig2}. Our results show that indeed, understanding the statistics of streams of English texts as a result of a grammatical ordering process (locally reshuffling), superimposed on a SSR word selection process, is consistent with the statistical evidence. 
While a pure SSR process offers a natural explanation for the observed (approximate) Zipf's laws in written texts, based on the necessity of contextual coherence, it fails to produce realistic word transition probabilities. Pure SSR transition probabilities are triangular and  very different from empirical transition probabilities. 
The natural  assumption that grammar is a process that locally rearranges word order, allows us to show that very simple grammatical rules are sufficient to explain the empirical structure of word transition matrices, in a statistical sense. 
Grammatical ordering that locally reshuffles selected words to comply with grammatical structures, sufficiently destroys the triangular (SSR) transition structure of the word selection process. goSRR models are therefore adequate, in the sense that they explain both the empirical word frequency distribution functions and basic statistical properties of the word transition probabilities of texts. 

Of course, we can not say that actual English language is a superposition of a SSR process 
and grammatical reordering. We have seen that the goSSR model, at the level of statistics of the transition probabilities, 
performs similarly well as the Bernoulli model, which we know is a truly bad model.  
However, the Bernoulli model can not explain the Zipf law in word frequencies. It can only explain features 
of the transition matrices, given that the that the Zipf law in word frequencies is provided as exogenous input. 

Note that this particular superposition of processes exemplifies a more general phenomenon. 
If highly structured processes interact with each other, the resulting (still very complex) process may 
look much more random than the underlying processes themselves. 
This emphasizes the often neglected fact, that statistical data alone is often insufficient for 
inferring the generative structure of the process that produces the data.
Only if a specific parametric process class can be identified as being adequate for describing a given 
phenomenon, then data can be used to estimate which process within that class is likely to have 
generated the data. 
In other words, in the context of entangled, possibly {\em multi-causal} generative processes, 
even ``big data'' becomes
worthless, without what is sometimes called a {\em thick description} of the phenomenon, 
which in mathematical terms, is the 
requirement of having identified the process class that produces the observed phenomenon reasonably well.  
A minimal way to ``thicken'' a description consists of considering a spectrum of measures 
that provides clues to the underlying structure of a process instead of reducing a 
complex phenomenon to a singular notion of similarity.

\section{Methods}
{\bf Implementing grammatical word order.}
We first identify the full-stop, exclamation mark, and question mark in an original text as sentence ends,  
and obtain the sentence lengths in the text body.
We produce a SSR  sequence, $x=(x_1,\cdots,x_N)$, of $N$ words, using a vocabulary of $W$ words.
The sequence is produced sentence by sentence, meaning that for every sentence in a text we generate a SSR sequence of the same length $m$ as the sentence $s=(s_1,\cdots,s_m)=(x_r,\cdots,x_{r+m-1})$, a sub-sequence of the text $x$, that starts at some position 
$r$ in the text. 
If the SSR process reaches word $i=1$ in mid-sentence, the SSR process is continued by restarting the SSR process. 
Each $x_t$ takes integer values between $1$ and $W$. 
For simplicity, assume here that there are no neutral words (grammatical label value $0$). 
To every of the $W$ distinct words $i=1,\cdots,W$, we randomly assign 
one of $N_g$ grammatical labels $L(i)$. 
We now work through $x$ sentence by sentence; the sentence length structure of the model
is defined by the sentence lengths in the original text that we model. 
 Let $s=(s_1,\cdots,s_m)$ be such a sentence.
Then we form sub-sequences of $s$ consisting only of words with a particular grammatical label, 
$L(s_t)=\ell$, where $1\leq t\leq m$, and 
$\ell=1,\cdots,N_g$. 
Let $t_n(\ell)$ be the index of the $n$'th word in the sentence $s$ that 
carries the grammatical label $\ell$, then $s_\ell=(s_{t_1(\ell)},s_{t_2(\ell)},\cdots,s_{t_{n_\ell}(\ell)}$, 
where $n_\ell$ is the number of words with label $\ell$ in sentence $s$. 
The sequences $s_\ell$  are typically not of the same length. To make all sequences equally long 
we define $s_{t_n}(\ell)$ to be an ``empty word'', whenever $n>n_\ell$. 
In this way we can think of the sequences, $s_{t_n}(\ell)$, to be all of the same length 
$\hat n=\max\{n_\ell|1\leq\ell\leq N_g\}$ and parse them in lexicographical order with respect to $(t,\ell)$,
where $(t',\ell')>(t,\ell)$, if $t'>t$, or $t'=t$ and $\ell'>\ell$.

The resulting sequence is the grammatically ordered sentence.
Note that the grammatically ordered SSR sequence and the SSR sequence $x$ have identical word frequency distributions.
If the SSR model explains Zipf's law, then so does the goSSR model. 
However, unlike SSR models, goSSR models now exhibit word transition probabilities 
that for a low enough fraction of neutral words (sufficiently small  $p_n$), and a complex grammar 
(sufficiently large $N_g$), statistical properties of the transition probabilities of the goSSR model start to resemble 
those of real texts.

For example, with $N_g=3$ classes, $1\equiv S$, $2\equiv P$, and $3\equiv O$, 
for each sentence we get three sub-sequences, $s_S$, $s_P$, and $s_O$. 
$s_{S}=(s_{t_1(S)},s_{t_2(S)},\cdots,s_{t_{n_S}(S)})$ is the 
sub-sequence of all words in $s$ carrying label $S$. It contains all $n_S$ words of the sentence that carry the label $S$.
We write $s_S(\tau)=s(t_\tau(S))$ for $\tau=1,\cdots,n_S$. 
Similarly, $x_P$ and $x_O$ are the sub-sequences for labels $P$ and $O$, respectively. 
After following the procedure described above, we obtain the sequence 
\begin{equation} 
(s_S(1),s_P(1),s_O(1),s_S(2),s_P(2),s_O(2),\cdots) \quad .
\label{exX}
\end{equation} 
Finally, by deleting the ``empty words'' we obtain a sentence that consists of the same words and has the same length as the SSR 
generated sentence $s$. This sentence is the grammatically ordered sentence.

If neutral words are present we proceed by combining neutral words with the next non-neutral 
word in a the sentence. For instance, if we find a sentence fragment of the form $\cdots,i,j,k,r,\cdots $ with grammatical labels 
$\cdots,3,0,0,2,\cdots $ then we consider $j,k,r$ as a single word $jkr$ with grammatical label $L(r)=2$.
After this blocking we proceed as before.
\\

{\bf Word transitions of texts and models.}
How similar are word transition properties of the goSSR model to those of actual texts?
For computational reasons we restrict ourselves to relative word transition frequency
matrices $A$ for the $W_{\max}=500$ most frequent words in a text. Since matrices for actual text, $A^{\rm text}$, 
can not be directly compared to those of a model, such as  $A^{\rm goSRR}$, we consider two appropriate statistical measures.  

The first is a proxy for the largest eigenvalue of $A$, the cosine of the angle between the vector $v=(1,1,\cdots,1)$ and the 
vector $Av$,  
\begin{equation}
		\cos(A)=\frac{(v|Av)}{|v||Av|}\quad.
	\label{equ:cosstat}
\end{equation}
It measures how quickly the transition probabilities transform an equi-distributed set of words into the 
stationary empirical word frequency distribution.

The second is a measure for the asymmetry, the skew of $A$, 
\begin{equation}
		{\rm skew}(A)\equiv\sum_{i=1}^{W_{\max}}\sum_{j=i}^{W_{\max}-1}(A_{ij}-A_{ji})\quad.
	\label{equ:skew}
\end{equation}
\\

{\bf Bernoulli model.}
To keep the same sentence structure as in the original text,  
the Bernoulli model is obtained by first locating the positions of sentence-ends in the text. 
Then remove the sentence-ends and randomly re-shuffle the words of the entire text. 
Finally, we reinsert sentence ends at the previous positions in the text. 
We reshuffle the words of the text while keeping the lengthy of the sentences fixed. 
\\

{\bf Preferential Simon model.}
To generate a Simon model to fit a text with a vocabulary of $W$ words, and length $N$ 
we 
propose 
the following version of the Simon model. 
We initialize the process with a vocabulary of 
$W_0$ words with initial weights $k_i(t=0)=k_0$, 
for $i=1,\cdots, W_0$, and $k_i(t=0)=0$, for all other $i$. 
We use $k_0=1$.
The probability $p_+(t)$ for sampling a new word at time step $t$ is 
computed from the size of the used vocabulary up to $t$, $W_{t-1}$, 
and the remaining number of time steps, $T+1-t$,
\[p_+(t)=\frac{W-W_{t-1}}{T+1-t} \quad.\]    
The probability of sampling word $i$ at time $t$ is given by 
\[p_i(t)=(1-p_+(t))\frac{k_i(t)}{\sum_{j=1}^W k_j(t)} \quad . \]    
Every time a word $i$ is sampled at time $t$, we increase $k_i(t+1)=k_i(t)+1$.
In this way the process follows a Simon type of update rules, while adapting  
its parameters to match length and vocabulary of a given text.
\\

{\bf Random typewriting model.}
Assume we have a keyboard (alphabet) with $V$ letters and a space key. 
The probability to hit the space key is $p_w(t)$. 
We initialize the model with an empty lexicon. 
For each time step $t=1,\cdots, N$, we sample a random sequence of letters (words), where each letter is produced  
with probability $(1-p_w(t))/V$. If a space is hit, sampling letters stops, 
and it is checked, if the random word already exists in the lexicon. 
If not, the word is added to the lexicon, and its word-count, $k_{\rm word}$, is set to one; 
otherwise, the word count is increased by one.
Knowing the number of words sampled up to time $t$, $W(t)$, and the number of time steps $T+1-t$ to go, 
one can adjust $p_w$ to control the total number of words sampled by random typewriting. 
Heuristic considerations suggest for example to choose $p_w(t)=1/(1+q(t))$, with
$q(t)=\log((W-W(t)+1)(V-1)+1)/\log(V)-1$.
For the random typewriting process text length and vocabulary size are harder to be matched with real texts.  
\\

{\bf Acknowledgements}\\
This work was supported by the Austrian Science Fund FWF under P29032 and P29252.
We gladly acknowledge important and insightful discussions on the presented topic with our dear colleague 
Bernat Cormominas-Murtra.

\end{document}